\documentclass{article}
\usepackage{spconf,amsmath,graphicx}

\title{Event-based Action Recognition Using Timestamp Image Encoding Network }
\name{Chaoxing Huang\thanks{This work was done when Chaoxing was at ANU}}
\address{The Australian National University\\chaoxing.huang@anu.edu.au}
\begin{document}
%
\maketitle
\begin{abstract}
Event camera is an asynchronous, high frequency vision sensor with low power consumption, which is suitable for human action recognition task. It is vital to encode the spatial-temporal information of event data properly and use standard computer vision tool to learn from the data. In this work, we propose a timestamp image encoding 2D network, which takes the encoded spatial-temporal images of the event data as input and output the action label. Experiment results show that our method can achieve the same level of performance as those RGB-based benchmarks on real world action recognition, and also achieve the SOTA result on gesture recognition. 
\end{abstract}
\begin{keywords}
Event-camera, Action recognition, Spatial-temporal encoding
\end{keywords}
\section{Introduction}
\label{sec:intro}
Action recognition plays significant roles in multimedia and robotic vision, such as sports video classification \cite{1}, autonomous driving \cite{2}, human robot interaction \cite{3} and safety surveillance \cite{4}. Conventional vision sensor in action recognition is camera, which suffers from problems of information redundancy, power consuming and motion blurry \cite{6}.  Event-camera, an asynchronous vision sensor, has the advantages of high frame-rated, power saving  and only records moving edges in the environment \cite{7}. Therefore, it is suitable for those vision tasks including dynamical information. For example, human action recognition. However, the data representation of event-based data is different from images, and we need to find a way of bridging the gap between event camera and those powerful learning-based computer vision tools (deep conventional network). It is also important for us to utilise useful temporal information in the event-data to benefit action recognition, since temporal data  contains motion-relevant information. These two problems are the motivation of learning-based action recognition of using event-cameras.

 \textbf{Relations to prior works}: Few works have connected deep learning method with event-based human action recognition. Amir et al. \cite{12} use event image to encode the spatial information into an image and apply it to the 2D CNN for hand-gesture recognition,which ignores the temporal information. Wang et al. \cite{29} encode event data point into spatial temporal point-cloud, while it does not reflect the polarity of the data. 3D CNN is used in \cite{41} to perform action recognition, which introduce a large number of parameters. The work in \cite{20} uses timestamp image to encode the spatial-temporal information in event data for human gait recognition, which is similar to our method.  However, they do not consider the polarity of the event. Besides, the model in those work are only applied to simple and single scenario in action recognition like hand gesture recognition and gait recognition.

To echo the motivation and the aforementioned problems, we propose a method of using timestamp image encoding to encode the spatial-temporal information from the event data sequence into frame-based representation, and use standard 2D CNN to classify the action label of the observed event sequence. The contributions of this paper are: \textbf{1)} We propose a simple but compact CNN-based method to perform event-based action recognition by using the timestamp image to encode spatial-temporal information.\textbf{2)} We compare the performance of applying purely spatial encoding (event image representation) as well as spatial-temporal encoding (timestamp image representation), and demonstrate that temporal information is indeed useful in event-based action recognition. \textbf{3)} We demonstrate that our method can achieve a competitive performance result on real world action scenario compared with those RGB-based method, even without using the RNN/LSTM architecture. Our method also surpass state of the art in gesture recognition task.

\section{Methodology}
\label{sec:pagestyle}
\subsection{Event to Frame Transformation}
We first briefly review the the model of event-camera \cite{10}. Let us define the intensity of brightness of an independent pixel as $I$, and what the hardware of the photoelectric system precepts is the log photocurrent, which is denoted as $L=\log(I)$(We refer it as “intensity” in the following context). We also define the information that the pixel collects as "event", which is denoted as $e=((x,y),t,p)$. $(x,y)$ is the position coordinate of the pixel in the sensing matrix, $t$ is the timestamp when the event is logged by the pixel and $p$ is the polarity, which indicates if the intensity at this pixel  becomes brighter or darker. The intensity change is determined by the hardware threshold. The polarity can be further expressed as $p=sign(\frac{d}{dt}\log(I_t(x,y)))$, where the output of $sign$ function is $+1$ or $-1$. In reality, multiple pixels in the sensing matrix record their events, and we get sequence of event data. 

 To bridge the gap between CNN and event data, two methods of projecting events into frame-based representation are presented below,namely timestamp-image and event-image.
 \subsubsection{Timestamp image}
 Timestamp image mainly encodes the time information of the event data into image form. Given a sequence of event data, we consider using a sliding-window with fixed time-length to accumulate the events until the difference between the latest timestamp value and the earliest timestamp value equals to the window length. The pixel value of every pixel is determined by the relative time value within the time-window. The mathematical model is given as below:\\
 Without lost of generality, we assume  $n$ events in  pixel $(x,y)$ take place within the time-window, which are $[e_1(x,y,t_1,p_1),\\ \cdots,e_n(x,y,t_n,p_1)]$ (ascending sorted according to time). We also have the latest and earliest timestamp values as $t_{end}$ and $t_{begin}$ within the time-window by taking all the events into consideration. The pixel value at $(x,y)$ can then be expressed as:
\begin{equation}
I_{x,y}=\frac{t_n-t_{begin}}{t_{end}-t_{begin}}
\end{equation}
Equation (1) indicates the latest relevant time of the event takes place in the given time-window.
\subsubsection{Event image}
The event-image is also called 2D-histogram, since it encodes the amount of event that take place in a certain pixel into the corresponding pixel value. The pixel value at $(x,y)$ can then be expressed as:\\
\begin{equation}
    I_{x,y}=\sum_{t \in T}\delta(x_k=x,y_k=y)
\end{equation}
where $T$ is the length of the time-window. Equation (2) indicates counting the number of event at every pixel.
\subsubsection{Encode the polarity}
Polarity value(+1 or -1) dose not appear in the above two equations, but this information cannot be simply ignored in event-based vision.To do that, we simply apply the timestamp image and event image transformation to both positive and negative event separately, which means  two frame-representations will be acquired, namely $I_p$ and $I_n$. We then merge these two frames into one frame representation as in \cite{26}, where all the pixel values are re-scaled to the range from 0 to 255 after merging. The visualisation of both representations are shown in Fig. 1

\begin{figure}[t]

\centering
\includegraphics[scale=0.35]{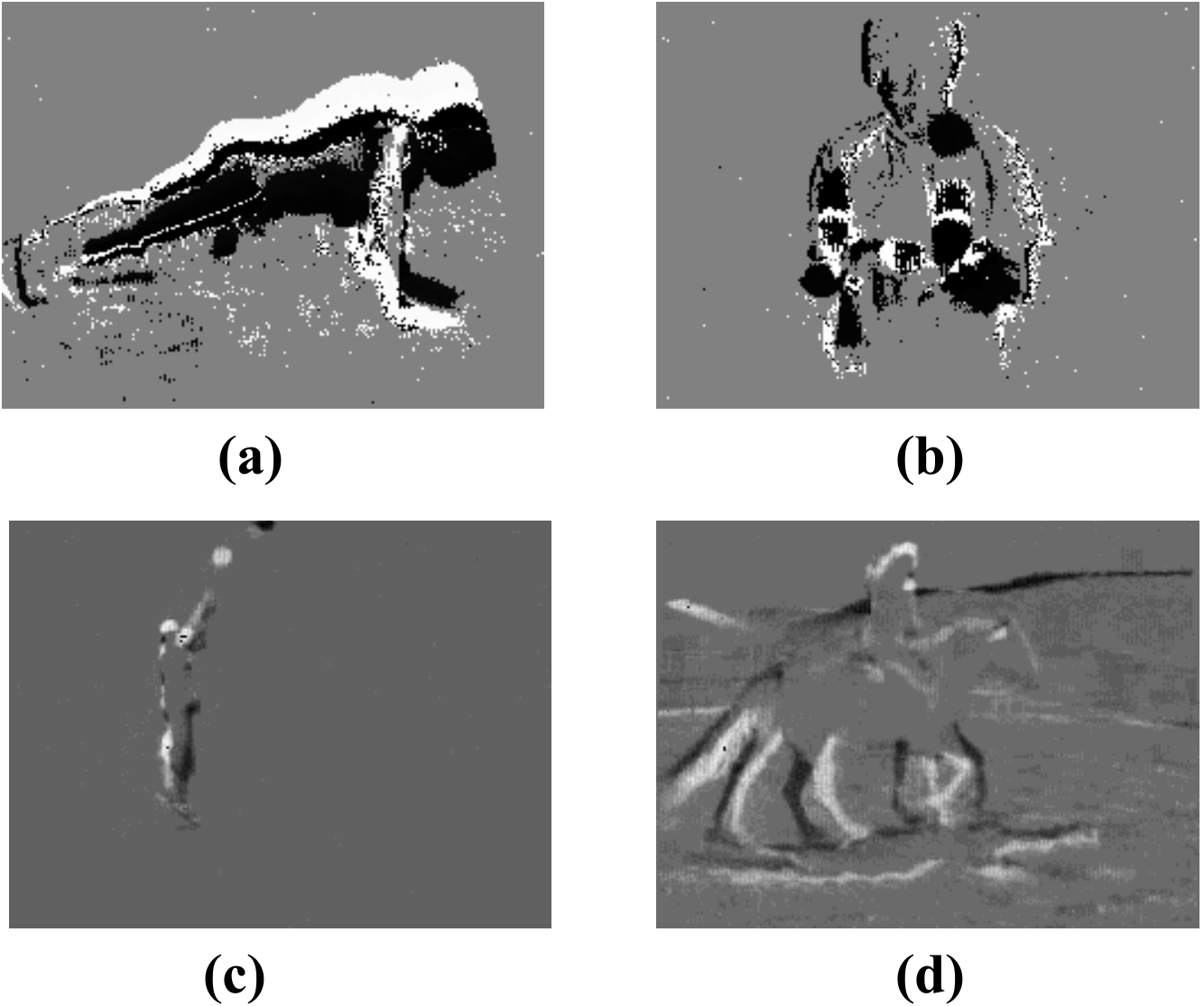}
\caption[Examples of timestamp image and event image]{Examples of timestamp image and event image.(a) Timestamp image of push up.(b) Timestamp image of ball juggling. (c) Event image of basketball throwing. (d) Event image of horse-riding. }
\label{fig:label}
\end{figure}

\subsection{Network Architecture and Pipeline}
\subsubsection{Timestamp-image chunk classifier}
We transform the event data into frame representation in the previous section, and we can now use CNN to apply recognition task to image data. One might consider training a classifier to recognize every frame, but it is problematic that some of the frames record very sparse pixel value since event camera is only sensitive to those moving objects. Those frames tend to have the moving background and some of the noisy pixels, which yields difficulties in recognizing human and other objects in the scene. If we directly train the classifier on a single frame, those noisy frame will create severe over-fitting problem and the classifier might even learn wrong features. To overcome this problem, we use a sliding frame-buffer with a step size of one frame, and it stores three event frame representation at every step. The stored three frames is referred as timestamp-image chunk, as we use timestamp-image representation in our model. At this stage, we train a classifier to classify the action label of every chunk. In other words, the classifier is trained to classify every single frame by looking two frames backward. The process of forming timestamp-image chunk is shown in Fig. 2. Following the network architecture in \cite{64}, we use the Pytorch version Resnext50 \cite{66} with pre-trained weight on Image-net as our classifier.

\begin{figure}[t]

\centering
\includegraphics[scale=0.9]{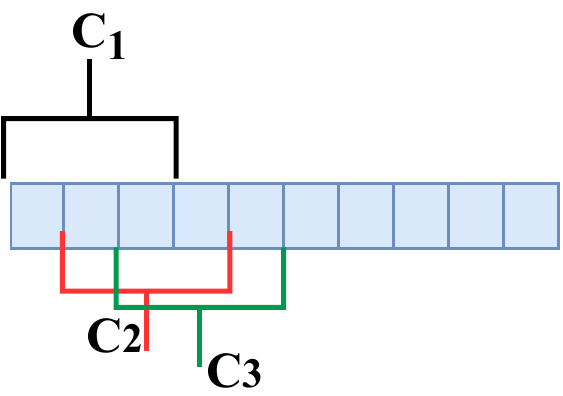}
\caption[Timestamp image chunk]{Process of creating timestamp image chunk. The buffer rolls one frame ahead as the video process.}
\label{fig:label}
\end{figure}
\subsubsection{Inference}
\par{When inferring the action label of an event video which has been transformed into frame representation, we will first use the sliding buffer to create chunks. Every chunk will be sent into the trained chunk-classifier and a class score vector will be obtained for every chunk.  Finally, temporal average pooling is applied to all the score vectors to get the final class score vector of the frame-based event video. }

\section{Experiments and results}
In this section, we demonstrate experiments on two challenge data-set. We will first introduce these two datasets. 
\subsection{Datasets}
 \textbf{DVS-128 gesture dataset } is an event-based human gesture dataset which is produced by IBM to build a low power cost event-based gesture recognition hardware system \cite{12}.  It was recorded by using the iniVation DVS-128 dynamic vision sensor to record a set of 11 hand and arm gestures from 29 subjects with 122 trails under 3 various lighting conditions. The number of the total instances is 1,342.  In our experiment, the gesture of "Other" is omitted and only 10 of the pre-designed gestures are included.
 
 \textbf{Event-version UCF-11 dataset} is recorded by playing the RGB dataset UCF-11 video on a computer monitor and a DAViS240C vision is put in front of the screen to get the event-data. The position of the vision sensor is adjusted to ensure the field of view can cover the region of interest on the monitor, and the video sequence is re-scaled to fit the field of view of the vision sensor. The video numbers and the categories remain the same as the RGB version apart from changing the data into event data as AEDAT form. The action samples in the dataset are all from real world videos. We refer the readers to the RGB version dataset for more details \cite{77,37}. Note that we do not use the UCF-50 dataset due to the 25-fold validation is computationally consuming, while the event-version UCF-101 is not published. \subsection{Implementation details}
 The hardware platform are  4 NVIDIA Titan-X GPUs and the i9-9900 CPU. The experiment environment is Ubuntu 16.04 and the framework is Pytorch 1.1.0 \cite{70}. The time-window of collecting event data for generating each frame representation is set as 80 ms and each frame is resized as 224 $\times$ 224. Random flipping and brightness jittering are used for data augmentation. As for the training of the classifier, the optimizer is Adam with $\beta_1=0.9$ and $\beta_2=0.999$. The learning rate and batch size are $1e-4$ and 64 respectively and a regularization of $1e^{-5}$ is used for prevent over-fitting.
 \subsection{Results}
 \subsubsection{Comparison on different event-representations}
We first compare the performance of using timestamp image chunk and  event-image chunk. For simplicity here, we group the 25 groups of data into 5 larger groups and do  5-fold validation for both representations respectively. This improves the difficulty  level of the recognition task since the number of learning sample is considerably reduced compared  with the original leave-one out validation(LOOV). Each fold of training takes 20 epochs. The results are shown in Table 1. As is shown, the performance of using timestamp-image chunk is higher than that of using event-image chunk. It indicates the effectiveness of applying spatial-temporal encoding to the event data, where the timestamp image has temporal information while the event image does not.  We use the former as the representation in the rest of our experiment.
 \begin{table}
\caption{Performance comparison of two different representations }
  \setlength{\tabcolsep}{10mm}{
  \begin{tabular}{||c c ||} 
 \hline
 \textbf{Representation} & \textbf{Accuracy(\%)}  \\ [0.5ex] 
 \hline\hline
 Timestamp-chunk & 83.23 \\ 
 \hline
 Event-chunk & 82.00 \\[1ex] 
 \hline
\end{tabular}
}
\end{table}
\subsubsection{Results on EV-UCF-11 dataset }
We conduct the experiment on human action recognition by using the EV-UCF-11 dataset, and we strictly follow the LOOV method to test the performance for consistency because we need to compare it with those benchmarks which use 25-fold validation. Benchmarks that use EV-UCF-11 as well as UCF-11(RGB) dataset are included in the comparison. The results are shown in Table 2. The data indicates that the proposed method outperforms the existing event-version benchmark(Motion map + MBH) significantly and reach the same performance level as those of RGB benchmarks. The results have demonstrated the effectiveness of our proposed algorithm since event data does not have much background information.  It should be noted that although the result of our method still falls behind that of the Two-stream LSTM method, it potentially uses less information.

\begin{table}
\caption{UCF-11 results comparison}
    \setlength{\tabcolsep}{8mm}{
 \begin{tabular}{||c c ||} 
 \hline
 \textbf{Method} & \textbf{Accuracy(\%)}  \\ [0.5ex] 
 \hline\hline
 Motion map + MBH\cite{71} & 75.13 \\ 
 \hline
 Two-stream LSTM \cite{73}& 94.60\\
 \hline
 Differential motion\cite{74} & 90.24\\
 \hline
 CNN+TR-LSTM\cite{75} & 93.80\\
 \hline
 CNN+LSTM\cite{75} & 92.30\\
 \hline
 \textbf{ Proposed} & \textbf{92.90}\\[1ex] 
 \hline
\end{tabular}
}
\end{table}
\subsubsection{Results on DVS-128 gesture dataset}
We also conduct gesture recognition experiments by using the DVS-128 gesture dataset and achieve the state of the art result. The results are shown in Table 3 . We also try to use a smaller or larger time-window size for event data collection, which turns out to have a lower performance.
\begin{table}[t]
\caption{DVS-128 results comparison}
\setlength{\tabcolsep}{8mm}{    
 \begin{tabular}{||c c ||} 
 \hline
 \textbf{Method} & \textbf{Accuracy(\%)}  \\ [0.5ex] 
 \hline\hline
 Pointnet ++ \cite{29} & 97.06 \\ 
 \hline
 Time cascade \cite{12} & 96.49\\
 \hline
 
 \textbf{Proposed, T=80ms} & \textbf{97.35}\\
 \hline
 \textbf{ Proposed,T=50ms} & \ 95.83\\
 \hline
 \textbf{Proposed, T=20ms} & \ 95.83\\[1ex] 
 \hline
\end{tabular}
}
\end{table}

\subsubsection{Insight of motion information of event-based vision}
One might be curious why does the event-based algorithm can achieve a similar performance level to those of RGB-based method even if the background, color and other contextual information is absent. We believe that witnessing the complete moving process of those edges is vital for making the right prediction. We apply a simple comparison to demonstrate this. Following the split-1 protocol in \cite{17}, we compare the action recognition accuracy of our algorithm and the RGB-baseline \cite{17} at  observation ratio of 10\% (early stage of the action). The results are shown in Table 4.
It turns out that event-based algorithm perform a much lower accuracy when the actions are incomplete, since motion information are not significant during the early stage.  

\begin{table}
\caption{ Performance  at  an observation ratio of 10\%}
    \centering
    \setlength{\tabcolsep}{5mm}{
 \begin{tabular}{||c c c ||} 
 \hline
  & RGB-based\cite{17} & Proposed  \\ [0.5ex] 
 \hline
  Accuracy & 89.57\% &79.78\%  \\ 
 \hline 
 
\end{tabular}
}
\end{table}
We demonstrate some of the early frame representations of event-based vision and RGB-based vision in Fig .3 , one(human) might find it rather hard to know what the human in the video is doing by only observe the frame at early stage without taking a look at the RGB frame. During the early-stage, background and contextual information becomes crucial while motion information is not enough. 
\begin{figure}[t]
\centering
\includegraphics[scale=0.4]{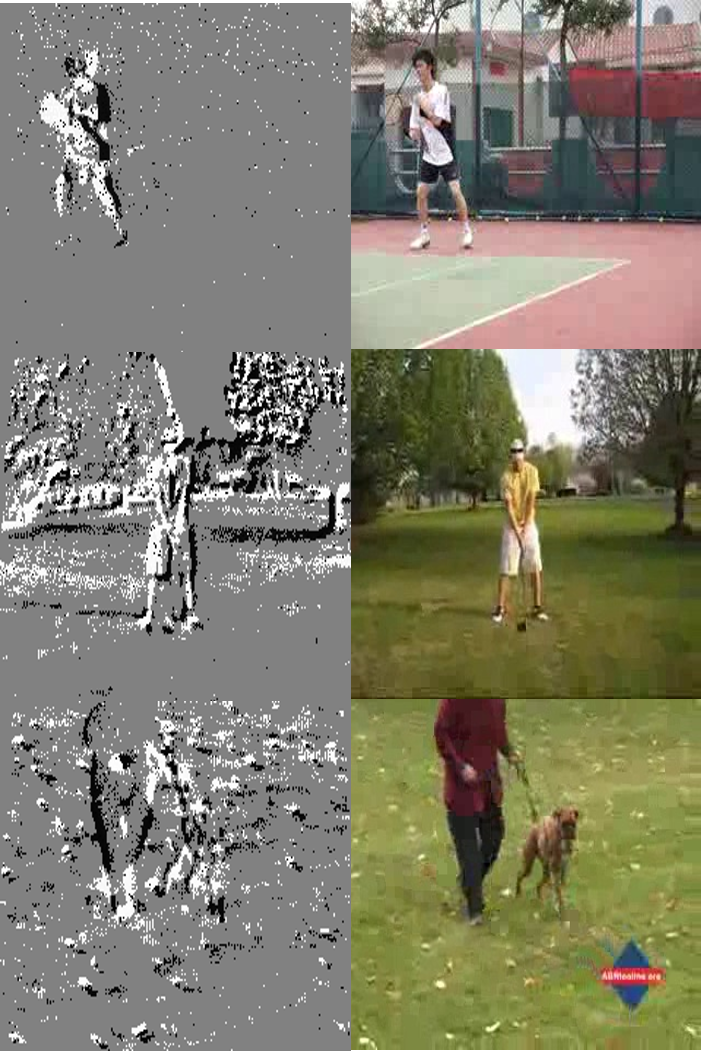}
\caption[Visualisation of hard examples]{Frame comparison at an observation ratio of 10\%. Left column: Timestamp images. Right column: RGB frames}
\label{fig:label}
\end{figure}
\section{Conclusion}
In this paper, we demonstrate a simple yet compact CNN-based method for event-based action recognition using timestamp image encoding. Experiment results show that our applying spatial-temporal encoding image can effectively use the temporal and motion information in action recognition. Our event-based method can be as competitive as its RGB-based counter part in real-world action recognition and it also achieve the state of the art result on gesture recognition. 
\section{Acknowledgement}
I would like to thank Dr Miaomiao Liu for providing useful advice and GPUs for this project. 



\bibliographystyle{IEEEbib}
\bibliography{strings,refs}

\end{document}